\documentclass{article}
\usepackage[final]{corl_2018}

\usepackage{graphicx}
\usepackage{amsmath,amssymb} 
\usepackage{color}
\usepackage{bbm}
\usepackage{comment}
\usepackage{subcaption}

\usepackage{algorithm,algorithmic}

\title{Model Learning for Look-ahead Exploration in  Continuous Control}

\author{Arpit Agarwal, Katharina Muelling and Katerina Fragkiadaki\\
Carnegie Mellon University\\
United States\\
\{arpita1,katharam\}@andrew.cmu.edu, katef@cs.cmu.edu
}

\begin{document}
\maketitle

\begin{abstract}
    We propose an exploration method  that incorporates look-ahead search  over basic learnt skills and their dynamics, and use it for reinforcement learning (RL) of manipulation policies .
    Our skills are  multi-goal policies learned in isolation in simpler environments using existing multigoal RL formulations, analogous to options or macroactions. Coarse skill dynamics, i.e., the state transition caused by a (complete) skill  execution, are learnt and are unrolled forward during lookahead search. 
    Policy search benefits from temporal abstraction during exploration, though itself operates over low-level primitive actions, and thus the resulting policies does not  suffer from suboptimality and inflexibility caused by coarse  skill chaining. We show that the proposed exploration strategy results in effective  learning of complex manipulation policies faster than current state-of-the-art RL methods, and  converges to better policies than methods that use options or parametrized skills as building blocks of the policy itself, as opposed to guiding exploration. We show that the proposed exploration strategy results in effective  learning of complex manipulation policies faster than current state-of-the-art RL methods, and  converges to better policies than methods that use options or parameterized skills as building blocks of the policy itself, as opposed to guiding exploration.
\end{abstract}

\section{Introduction}

In animals, skill composition greatly increases the efficiency to solve new problems \cite{competence}.  
The skills act as the “building blocks” out of which an agent can form solutions to new problem configurations as well as entirely new problems \cite{NIPS2004_2552}. 
Instead of creating a solution from low-level motor commands for each new challenge, skill composition enables the agent to focus on combining and adjusting higher-level skills to achieve the goal. This principle of hierarchical skill composition has been applied by researchers in Reinforcement Learning (RL) in the hope to achieve a similar efficiency for artificial agents \cite{SUTTON1999181,DBLP:journals/corr/KulkarniWKT15}. 
In this context, skills are often referred to as options or macro-actions \cite{SUTTON1999181}. Options realize the idea of temporally extended actions that can independently accomplish a sub-goal for a defined set of scenarios.  
The higher-level policy is then tasked with obtaining the optimal sequence of options to accomplish the task. The performance of the policy therefore critically depends on the set of options available to the agent. 
If the option set is poorly chosen, the resulting composed policies will be suboptimal  \cite{DBLP:journals/corr/cs-LG-9905014}. Many researchers have tried to  find a ``golden" set of options \cite{NIPS2004_2552} to compose hierarchical policies from \cite{DBLP:journals/corr/cs-LG-9905014}. 
However, with a growing number of options the efficiency of learning will suffer. This leads to a \textbf{trade-off between flexibility and learning speed}: the fewer options, the faster the learning, the less optimal the resulting composed policy. 
In light of the above, this paper proposes  temporally abstracted look-ahead search for exploration, yet fine-grained action composition for  policy representations in the manipulation domain. In other words, a small set of generalized basic manipulation policies, which we call skills or options, are learnt and their coarse transition functions are used to unroll forward a tree search during the exploration loop. Yet,  policy search still operates over low-level primitive actions, and thus the resulting policies are  not  limited to coarse skill  compositions, as previous hierarchical reinforcement learning formulations \cite{SUTTON1999181}.    This design choice accelerates learning while at the same time permits  flexibility in option (skill) selection: as long as the set of skills and the states they visit sufficiently covers the  state space necessary for complex manipulation, skills can be redundant, overlapping, or varying in duration, without loss in performance of the final policy. 

\begin{figure}[t!]
    \centering
    \includegraphics[width=1.0\linewidth]{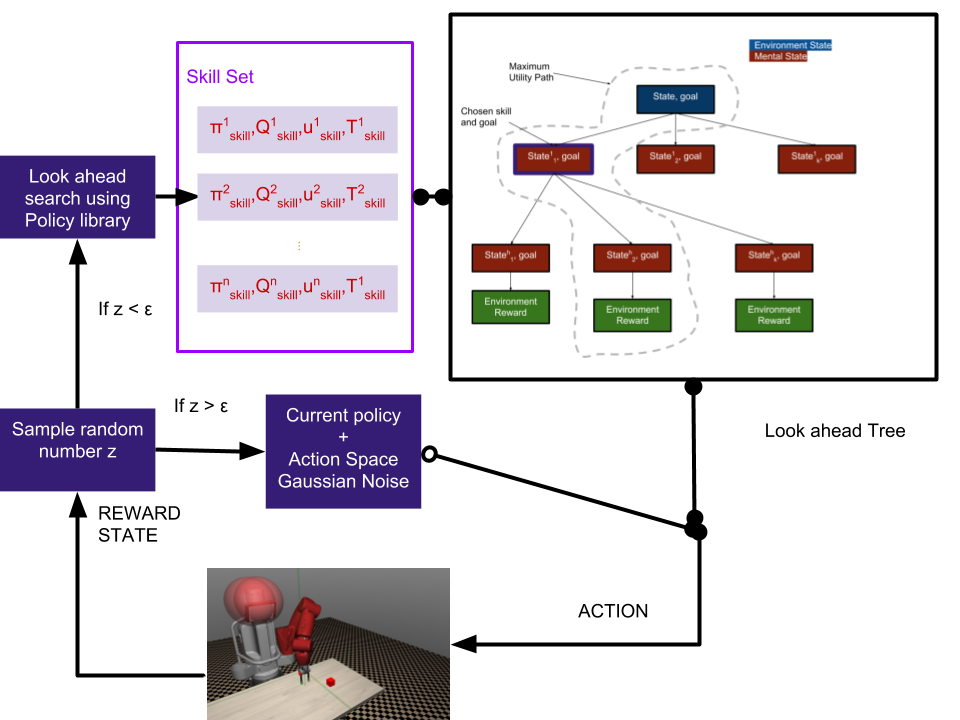}
     \centering
    \caption{\textbf{Overview}. We use learned skill dynamics with deep neural regressors and use them for look-ahead tree search,  to guide effective exploration in reinforcement learning of complex manipulation tasks.}
    \label{fig:overview}
\end{figure}

In a nutshell, our framework works as follows. We  train a set of simple manipulation skills, such as grasp, transfer, and reach.  
Each of these skills represents a (generalized) policy that can handle a set of related goals and is parametrized by both the current state and the desired goal (see Figure \ref{fig:overview}). Each skill may involve a different number of objects, thus   different skills may have different state representations. For each skill we learn a coarse-grain transition function, 
that, given a pair of initial and goal states, it predicts the resulting state, after execution of the skill. 

During  training of a new manipulation task represented by a goal configuration, we perform look-ahead tree search  using  the coarse-grain  learned skill dynamics: at each time step of an episode, we unfold a search tree by sampling \textit{skill-goal} combinations and using the learned neural skill dynamics to predict resulting states (Figure \ref{fig:overview}). We choose the tree path that leads closer to the desired  goal configuration, and the first skill-goal of the path is executed in the real environment. We execute the chosen skill with its goal parameter until termination of that skill or episode.
Then,  planning is repeated from the newly reached state, akin to model predictive control \cite{Mayne20142967}. 

 Our skill-based look-ahead exploration outperforms  $epsilon-$greedy exploration, model-based RL \cite{Sutton:1991:DIA:122344.122377} where the fitted dynamics model  is used to supply (fake) experience tuples and not  for exploration,  as well as  policy learning over coarse parameterized skills \cite{ICLR16-hausknecht}, as opposed to low-level action primitives.
 
In summary, success of our framework depends on the following design choices, that distinguish it from previous works:

\begin{itemize}
    \item  Look-ahead with coarse-grain skill dynamics yet act with fine-grain primitives. Coarse-grain  transition dynamics do not suffer from severe model error accumulation when unrolled in time \cite{DBLP:journals/corr/OhGLLS15}, as  few  hops are sufficient to look far into the future.
    Yet, policies over fine-grained actions --- as opposed to pretrained  skills ---   produce smooth  behaviour, being as fast and optimal as possible.
    \item Purposeful skill dynamics.  Our dynamics model used for look-ahead is built from basic manipulation skills, as opposed to  random exploration and transitions therein \cite{DBLP:journals/corr/abs-1803-10122}. This ensures that learnt dynamics cover important part of the state space. In contrast, dynamics learnt from random exploration alone often miss useful states, and thus their temporal unrolling is less informative.
\end{itemize}

To find out more about our work visit the   \href{https://sites.google.com/view/skill-based-exploration/home}{project page}. Our code is available at \href{https://github.com/arpit15/skill-based-exploration-drl}{github}.

\section{Related work}
\label{sec:related}
\paragraph{Exploration - Intrinsic motivation}
Effective exploration is a central challenge in learning good control
policies \cite{NIPS2015_5668}. Methods such as $\epsilon-$greedy, that either follow the current found policy or sample a random action with a certain probability, are useful for local exploration but fail to provide impetus
for the agent to explore different areas of the state space. Exploring by maximizing the agent's curiosity  as measured by the error of predictive dynamics \cite{DBLP:journals/corr/PathakAED17}  expected improvement of predictive dynamics \cite{Schmidhuber:1991:PIC:116517.116542},   information maximization \cite{NIPS2015_5668},  state visitation density \cite{DBLP:journals/corr/BellemareSOSSM16}, uncertainty of the  value function estimates \cite{DBLP:journals/corr/OsbandBPR16}, all have been found to outperform  $\epsilon-$greedy, but are limited due to the underlying models operating at the level of basic actions. In  manipulation, often times there are very few actions that would lead to effective new outcomes and are hard to discover from uninformative landscapes of the reward function.

\paragraph{Multigoal RL - Inverse dynamics}
The Horde architecture \cite{Sutton:2011:HSR:2031678.2031726}  proposed to represent  procedural knowledge in terms of multiple Value Functions (VF) learned in parallel by multiple RL agents, each with its own reward and termination condition. VF would capture time for specific events to happen, or, used for inducing a policy for a specific goal. 
Work of \cite{pmlr-v37-schaul15} proposed to represent a large set of optimal VF by a single unified function approximator, parametrized by both the state and goal. 
This takes advantage of the fact that similar goals can be achieved with similar policies, and thus allows generalization across goals, not only across states. 
Hindsight Experience Replay (HER) \cite{andrychowicz2017hindsight} introduces the simple idea that failed executions -episodes that do not achieve the desired goal- achieve some alternative goal, and is useful to book-keep in the experience buffer such ``failed" experience as successful experience for that alternative goal. Such state and goal parametrized experience is used  to train a generalized policy, where actor and critic networks take as input both the current state and goal (as opposed to the state alone). Thanks to the smoothness of actor and critic networks, achieved goals that are nearby the desired one, implicitly guide learning, instead of being discarded. Our approach builds upon HER, both for learning the basic skills, as well as the final complex manipulation policies, yet we propose novel exploration strategies, instead of $\epsilon$-greedy used in \cite{andrychowicz2017hindsight}.

Generalized (multigoal) policies learn to transition for a set of initial states to a set of related goals, and as such,  they are equivalent to \textit{multistep} inverse dynamics models. Many works in the literature attempt to learn inverse models with random exploration \cite{DBLP:journals/corr/AgrawalNAML16} and chain them in time for imitation \cite{DBLP:journals/corr/NairCAIAML17}. 
Our work  learns skill inverse models using multigoal RL and explicit rewards.  We are able to discover more useful and interesting temporally extended inverse dynamics models, which is unclear how to obtain with random exploration. For example, a robot would potentially never learn to grasp guided solely by random controls or curiosity, or at least, this result has not been obtained till today. 

\paragraph{Hierarchical RL}
Learning and operating over different levels of temporal abstraction is a key challenge in tasks involving long-range planning. In the context of reinforcement learning, \cite{SUTTON1999181} 
proposed the options framework, which involves abstractions over the space of actions. 
At each step, the agent chooses either an one-step primitive action or a multi-step action 
policy (option). Each option defines a policy over actions (either primitive or other options)
and can be terminated according to a stochastic function.  
 The MAXQ framework \cite{DBLP:journals/corr/cs-LG-9905014} 
decomposes the value function of a Markov Desicion Process (MDP) into combinations of value functions
of smaller constituent MDPs. 
Work of \cite{DBLP:journals/corr/KulkarniNST16NIPS2015_5668} learns a 
policy for scheduling semantically meaningful goals using deep Q networks. Work of \cite{learningbyplaying} uses simple intrinsic perceptual rewards to learn subtasks and their scheduling for helping learning of extrinsic motivated (non-hierarchical) policies. \cite{NIPS2004_2552} also explored agents with intrinsic reward structures in order to learn
generic options that can apply to a wide variety of tasks. Using a notion of salient events 
as sub-goals, the agent learns options to get to such events. Other works have proposed parametrized actions of discrete and continuous values as a form of macro-actions (temporally extended actions)  to choose from  \cite{ICLR16-hausknecht,DBLP:journals/corr/MassonK15}. We compare with the model from \cite{ICLR16-hausknecht} in the experimental section. Other related work for hierarchical formulations include Feudal RL  \cite{NIPS1992_714}
which consists of ``managers"  taking decisions at various levels of granularity, percolating all
the way down to atomic actions made by the agent.  \cite{daniel2016probabilistic} jointly  learn options and  hierarchical policies over them. Such joint search makes the problem more difficult to solve, moreover, options are not shared across policies of different tasks.  We instead capitalize over already known options (skills) to accelerate training of more complex ones.

\paragraph{Hierarchical Planning}
Planning has been used with known transition dynamic models of the environment to help search over optimal actions to take.  Incorporating macro-actions to reduce the computational cost of long-horizon plans has been explored in \cite{he2010puma}.  
In \cite{vien2015hierarchical}, the authors integrate a task hierarchy into  Monte Carlo Tree Search. These approaches work for discrete state, action and observation spaces, and under known dynamics.  
Our work instead considers continuous actions and states, and unknown dynamics.
\paragraph{Model-based RL}
To address the large sample complexity of model-free RL methods, researchers learn models of the domain, which they use to sample fake experience for policy learning 
\cite{Sutton:1991:DIA:122344.122377}, initialize a model-free method \cite{DBLP:journals/corr/abs-1708-02596} which is further fine-tuned with real experience to fight the biases of the model, or is combined with a model-free estimation of the residual errors \cite{DBLP:journals/corr/ChebotarHZSSL17}. When the model of the domain is given, Monte Carlo tree search has shown to be very effective for  exploration \cite{Silver2017MasteringTG}, and outperforms corresponding model-free RL methods \cite{NIPS2014_5421}, even when the latter are allowed to consume a great amount of (simulated) experience. 

\section{Exploration using look-ahead search over skill dynamics}
We consider a multi-goal Markov Decision Process (MDP) \cite{pmlr-v37-schaul15},
represented by states $\mathrm{s} \in \mathcal{S}$, goals $\mathrm{g} \in \mathcal{G}$, actions $\mathrm{a} \in \mathcal{A}$. At the start of each episode, a state-goal pair is sampled  from the initial state-goal distribution  $\rho(\mathrm{s}_0,\mathrm{g})$. 
Each goal $\mathrm{g}$ corresponds to a reward function  $r_{\mathrm{g}} : \mathcal{S} \times \mathcal{A} \rightarrow \mathbb{R}$. 
At each timestep, the agent gets as input the current state $\mathrm{s}_t$ and goal $\mathrm{g}$, and chooses an action $\mathrm{a}_t$ according to a policy $\pi: \mathcal{S} \times \mathcal{G} \rightarrow \mathcal{A}$,  and obtains reward $r_t = r_{\mathrm{g}}(\mathrm{s}_{t},\mathrm{a}_t)$. 
The objective of the agent is to
maximize the overall reward.
\par
\subsection{Learning multi-goal skill policies}
We  endow our agent with an initial set of  manipulation skills. We define a skill  as a short-horizon generalized action policy that achieves \textit{a set of related goals}, as opposed to a single goal. The goal sets of individual skills are not related to the goals of our final  manipulation tasks. 
Our framework can handle any set of skills, independent of their length, complexity, state and action spaces. Furthermore, skills can be redundant within the set. 
We  trained three skills: 
\begin{itemize}
    \item \textit{reach}, in which the gripper reaches a desired location in the workspace while not holding an object,
    \item \textit{grasp}, in which the gripper picks up an object and holds it at a particular height,
    \item \textit{transfer}, in which the gripper reaches a desired location in the workspace while  holding an object.
\end{itemize}
The skills do not share the same state space:  each involves different number of objects or it is oblivious to some part of the state space. 
Control is carried out in task space, by predicting directly $dx,dy,dz$ of the motion of the end-effector and the gripper opening. Task space control allows easier transfer from simulation to a real robotic platform and is agnostic to the exact details of the robot dynamics. The skills do not share the same action space either, e.g., the reaching skill does not control the gripper open/close motion. State and action abstraction allows faster skill training. Details on the particular skill environments and corresponding states, actions and rewards functions are included in the supplementary material. 

Each skill is trained using Hindsight Experience Replay \cite{andrychowicz2017hindsight} (HER) and off-policy deep deterministic policy gradients (DDPG)  \cite{DBLP:journals/corr/LillicrapHPHETS15} with standard  $\epsilon$-greedy exploration \cite{andrychowicz2017hindsight}. This allows us to decouple exploration and policy learning.
The agent maintains actor $\pi : \mathcal{S} \times \mathcal{G} \rightarrow \mathbb{A}$ and action-value (critic)  $\mathrm{Q} : \mathcal{S} \times \mathcal{G}\times \mathcal{A} \rightarrow \mathbb{R}$ function approximators. The actor is learned by taking gradients with respect to the  loss function $\mathcal{L}_a = - \mathbb{E}_s \mathrm{Q}(\mathrm{s},\mathrm{g},\pi(\mathrm{s},\mathrm{g}))$ and the critic minimizes TD-error using TD-target  $y_t = r_t + \gamma \mathrm{Q}(\mathrm{s}_{t+1}, \mathrm{g}, \pi(\mathrm{s}_{t+1},\mathrm{g}))$, where $\gamma$ is the reward discount factor. 
Similar to \cite{andrychowicz2017hindsight}, we use a binary reward function $r_{\mathrm{g}}$ which is 1 if the resulting state is within a specified radius  from the goal configuration $\mathrm{g}$, and 0 otherwise. 
Exploration is carried out by adding $\epsilon$ normal stochastic noise \cite{andrychowicz2017hindsight} to actions predicted by the current policy. 

HER, alongside the intended experience tuples of the form
$(\mathrm{s}_t,\mathrm{a}_t,r\mathrm{g}_t,\mathrm{g}, \mathrm{s}_{t+1})$, given the resulting state  $\mathrm{s}_T$ of an episode of length $T$, it adds additional experience tuples in the buffer, by considering $\mathrm{s}_T$ to be the intended goal for the experience collected during the episode, namely, adds tuples of the form $(\mathrm{s}_t,\mathrm{a}_t,r(\mathrm{s}_T)_t,\mathrm{s}_T, \mathrm{s}_{t+1})$. All the tuples in the experience buffer are use to train the actor and critic networks using the aforementioned reward functions. For more details, please refer to \cite{andrychowicz2017hindsight}.

\subsection{Learning coarse-grain skill dynamics and success probabilities} 
Multi-goal skill policies are used to obtain \textbf{general and purposeful forward dynamic models, that cover rich part of the state space and which are not easy to learn from random exploration. } 
For the $i$th skill,  we learn
\begin{itemize}
    \item 
a \textbf{coarse} transition function of the form $\mathcal{T}_{coarse}^i: (\mathrm{s},\mathrm{g})\rightarrow \mathrm{s}_{final}$ which maps an initial state $\mathrm{s}$ and goal $\mathrm{g}$ to a resulting final state $\mathrm{s}_{final}$, and
\item
a success probability function $u^i(\mathrm{s},\mathrm{g})\rightarrow [0,1]$, that maps the current state and goal to the probability that the skill will actually achieve the goal. 
\end{itemize}
We learn $\mathcal{T}_{coarse}$ and $u$ after each skill is trained. Data tuples are collected by sampling initial states and goals  and running the corresponding skill policy. The collected data is used to train  deep neural regressors for each skill, a three layer fully connected network that takes as input a state and a goal configurations and predicts the final state reached after skill execution, and the probability of success. The detailed architecture of the dynamics neural  networks is included in the supplementary.  Each manipulation skill is represented by   (generalized) policy $\pi$, action-value function $Q$, transition dynamics and probability of success: 
 $\mathcal{K}=\{ \Omega^i= (\pi^i, \mathrm{Q}^i, u^i, \mathcal{T}^i_{coarse}), i=1 \cdots N \}$, where $N$ is the number of skills and $\mathcal{K}$ the skill set.
 For us $N=3$.
 Our coarse skill dynamics model learns to predict the outcome of skill execution (on average 25 timesteps long) instead of predicting the outcome of low-level actions. This allows to plan over longer horizons without severe dynamics error  accumulation. 
Although HER with $\epsilon$-greedy exploration  successfully learns generalized basic skills guided solely  by binary reward functions, it fails to learn more complex skills, such as, \textit{put object A inside container B}, that require longer temporal horizon. 
Next, we describe how we use knowledge of coarse skill dynamics $\mathcal{T}_{coarse}$ and success probabilities $u$ for  exploration during training of such complex manipulation policies. 

\subsection{Exploration with look-ahead search}
We will use again DDPG with HER to train manipulation policies over low-level actions, but with a better exploration method. 
DDPG, being an off-policy algorithm, allows use to decouple learning and exploration. We exploit this decoupling and place our look-ahead search at the exploration step. 
With  $\epsilon$ probability we use the proposed look-ahead search to select 
the next action to try $\mathrm{a}_t$, otherwise we follow the current policy learned thus far $\pi(\mathrm{s}_t,g)$, where $\mathrm{s}_t$ denotes the current state. We vary $\epsilon$ to be close to 1 in the beginning of training, and linearly decay it to 0.001.  

\begin{figure}[h!]
    \centering
    \includegraphics[width=1.0\linewidth]{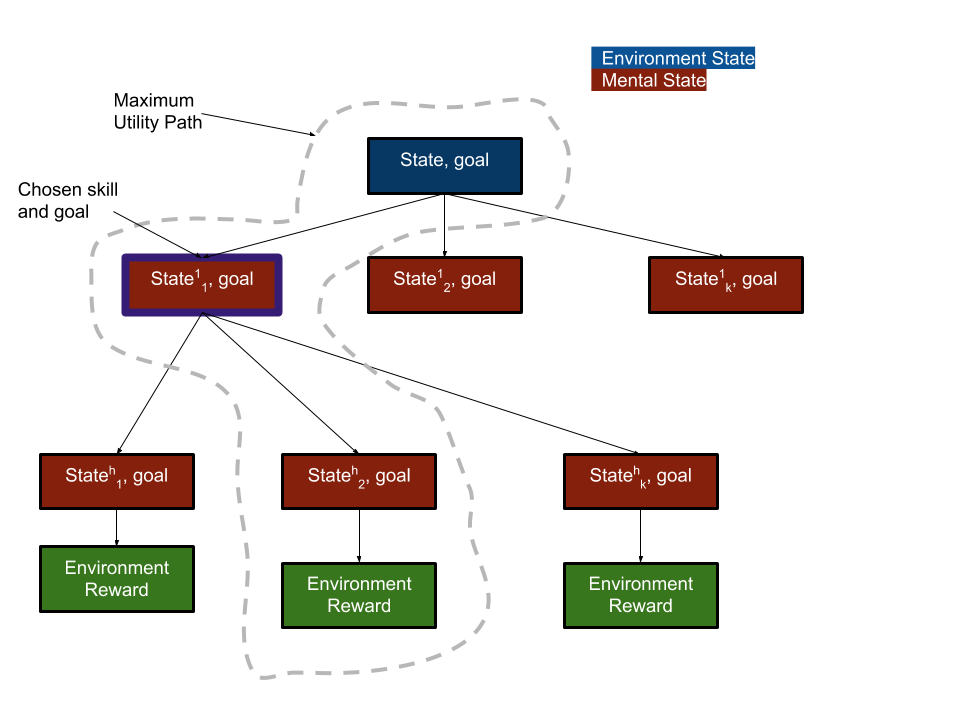}
     \centering
    \caption{\textbf{Look-ahead Search:} From the current environment state, we sample skill identities and skill sub-goals and (mentally) unfold a look-ahead tree using  learned skill dynamics. We select the first skill and sub-goal  of the path with maximum utility and  execute it in the ``real world".}
    \label{fig:lookaheadtree}
\end{figure}

Our look-ahead search works as follows: 
At each episode step, we unfold a search tree by sampling (with replacement) at each tree level, for each branch, a set of $K$ skill identities (in our case one of \textit{reach},  \textit{grasp} or \textit{transfer}),  and corresponding $K$  skill sub-goals, where $K$ is the branching factor of the search. For each sampled skill identity and sub-goal, we use the learned  success probabilities of each  sampled skill and sub-goal combination and prune improbable  transitions (line 13 in Algorithm \ref{fig:lookaheadtree}). For the remaining skill/subgoal combinations we (mentally) transition to the resulting final state  following  the learned   skill dynamics function.   
After  unfolding the tree for a prespecified number of steps, we choose the path with the maximum total reward defined as the sum of the transition rewards(reward for going from one node to other connnected node) of the skill executions as measured by the skill critic networks $Q^{i}$ and proximity of the final state to the desired goal configuration: $R = \sum_{(s, k \in \mathcal{K}, g_{k}) \in \text{path to leaf node}} Q^{k}(s,g_{k}) + r_{\text{final}}$, where $r_{\text{final}}$ is the negative of the Euclidean distance between the final state and the desired goal configuration $\mathrm{g}$. We execute the first (skill, sub-goal) tuple on the chosen maximum utility path in the current (``real") environment until the skill terminates, i.e., until the skill sub-goal is achieved  or maximum skill episode length is reached or goal is reached. The experience in terms of tuples $(\mathrm{s}_t,\mathrm{g},\mathrm{a}_t,r_t, \mathrm{s}_{t+1})$ populate the experience replay buffer. 

Note that the learned skill dynamics $\mathcal{T}_{coarse}$ and $u$ may not match the dynamics in the current environment. The reason could be both due to approximation error of the neural network regressors and the difference in the environment dynamics, e.g., task environments may contain additional objects on which the gripper can collide, which were not present during skill learning.   
Our look-ahead exploration is described in Algorithm \ref{algo:search} and visualized in Figure \ref{fig:lookaheadtree}. The complete exploration and reinforcement learning method is described in Algorithm \ref{algo:train}.


\begin{algorithm}[t]
\caption{HER with look-ahead search exploration(HERLASE)}\label{algo:train}
\begin{algorithmic}[1]
  \STATE \textbf{Input:} 
\STATE\hspace{\algorithmicindent} skill set $\mathcal{K}$
\STATE\hspace{\algorithmicindent} reward function $r: -\mathbbm{1}\lbrack f_g(s)=0\rbrack$
\STATE\hspace{\algorithmicindent} $\epsilon \leftarrow 1$, skill terminated $\leftarrow$ true
  \STATE Initialize $\pi, Q$, Replay buffer B
  \FOR{episode = 1, M}
    \STATE Sample a goal $g$ and starting state $s_0$
    \WHILE{episode not done}
        \IF{random(0,1) $< \epsilon$}
            \label{algoline:exploration}\IF{skill terminated}
                \STATE $\Omega^i, g^i$ $\leftarrow$ TreeSearch($s_t$, $g$)
                \STATE $( \pi^i, Q^{i}, u^{i}, \mathcal{T}^{ i}_{\text{coarse}}) \leftarrow \Omega^i$
            \ENDIF
            \STATE $a_t$ = $\pi^i(s_t, g^i)$
        \ELSE
            \STATE $a_t$ = $\pi(s_t,g)$ + Gaussian noise 
        \ENDIF \label{algoline:explorationend}
        \STATE $s_{t+1}$, $r_t$, terminal = execution($a_t$)
        \STATE skill terminated $\leftarrow$ checkSkillTermination($s_{t+1}$)
    \ENDWHILE
    \STATE Create hindsight experience with $g' = s_T$
  \ENDFOR
\end{algorithmic}
\end{algorithm}

\begin{algorithm}
\caption{TreeSearch}
\begin{algorithmic}[1]
    \STATE \textbf{Input:} 
    \STATE \hspace{\algorithmicindent}maxHeight H, branchingFactor B
    \STATE \hspace{\algorithmicindent}goal $g$, initial state $s_t^{\text{real}}$, skill set $\mathcal{K}$
  \STATE Initialize 
  \STATE\hspace{\algorithmicindent} root $\leftarrow$ ($s_t^{\text{real}}$,0, 0)
  \STATE\hspace{\algorithmicindent} openlist $\leftarrow$ addRoot
  \STATE\hspace{\algorithmicindent} leafNodelist $\leftarrow$ \{\}
  \WHILE{all path explored}
    \STATE $s$, currHeight = getTopLeafNode(openlist)
    \STATE sampled Set = sample skills and goal parameters(B)
    \FOR{$\Omega^i, g_i$  $\in$ sampled Set}
        \STATE $( \pi^i, Q^{i}, u^{i}, \mathcal{T}^{ i}_{\text{coarse}}) \leftarrow \Omega^i$
        \IF{$u^i(s,g_i) > 0.5$}
            \STATE nextState $\leftarrow \mathcal{T}_{coarse}^{i}(s, g_i)$
            \STATE $a_i  \leftarrow \pi^i(s,g_i)$
       
            \STATE transitionReward $\leftarrow Q^i(s, g_i, a_i)$
            
            \IF{currHeight+1 $<$ H}
                \STATE AddToLeafNodelist(nextState, \\
                            \hfill transitionReward, currHeight+1)
            \ELSE
                \STATE AddNodeToOpenlist(nextState, \\
                \hfill transitionReward, currHeight+1)
            \ENDIF
        \ENDIF
    \ENDFOR
  \ENDWHILE
  \STATE bestPath = getBestPath(leafNodeList)
  \STATE Return first skill and goal parameter of bestPath
\end{algorithmic}
\label{algo:search}
\end{algorithm}

\section{Experiments} \label{sec:experiments}
We test the proposed method in the MuJoCo simulation environment \cite{todorov2012mujoco} using a seven degree of freedom Baxter robot arm with parallel jaw grippers in the following  suite of   manipulation tasks: 

\begin{figure}
\centering
\begin{subfigure}{.49\textwidth}
  \centering
  \includegraphics[width=\linewidth]{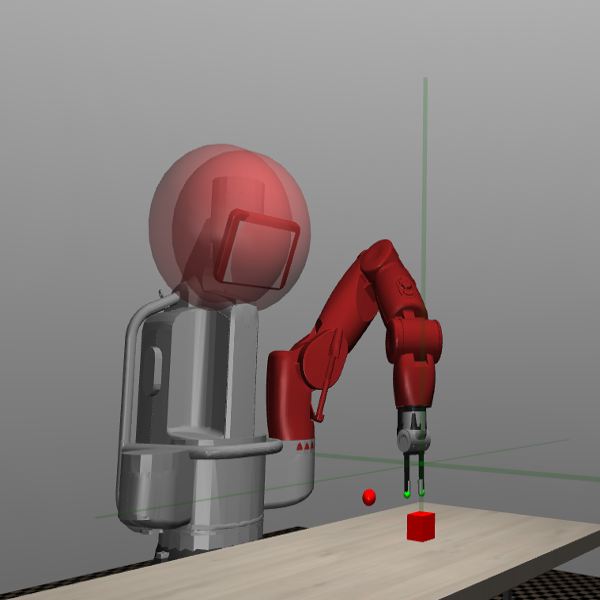}
  \caption{Pick and and Move}
  \label{fig:sub1}
\end{subfigure}
  \centering
\begin{subfigure}{.49\textwidth}
  \includegraphics[width=\linewidth]{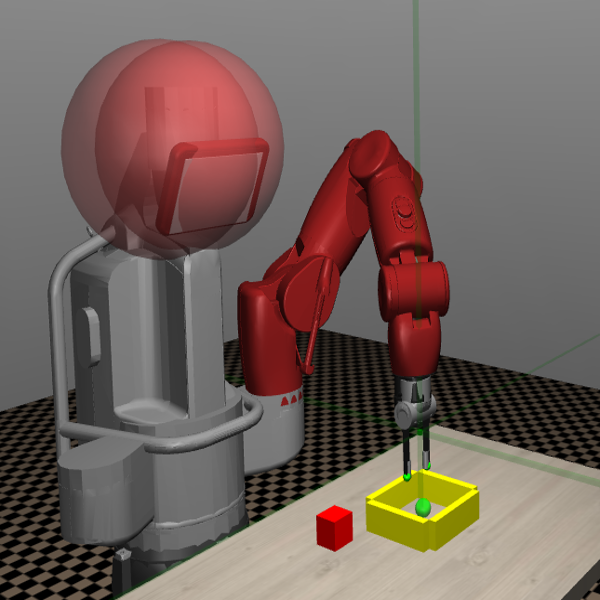}
  \caption{Put A inside B}
  \label{fig:sub2}
\end{subfigure}

\begin{subfigure}{.49\textwidth}
  \centering
  \includegraphics[width=\linewidth]{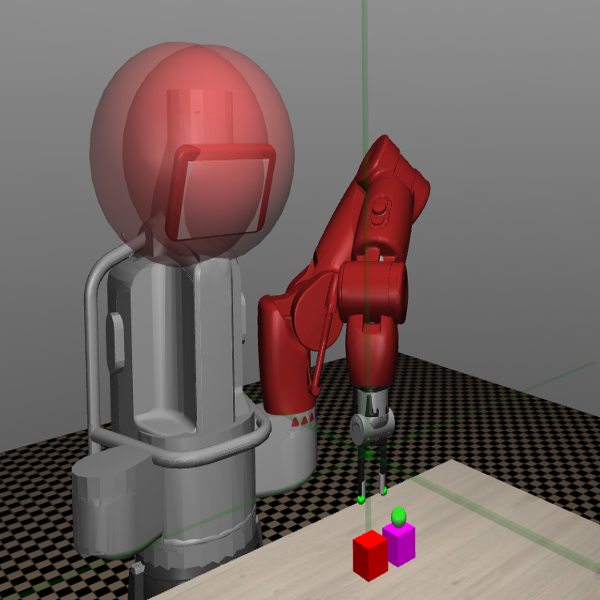}
  \caption{Stack A on top of B}
  \label{fig:sub3}
\end{subfigure}
\begin{subfigure}{.49\textwidth}
  \centering
  \includegraphics[width=\linewidth]{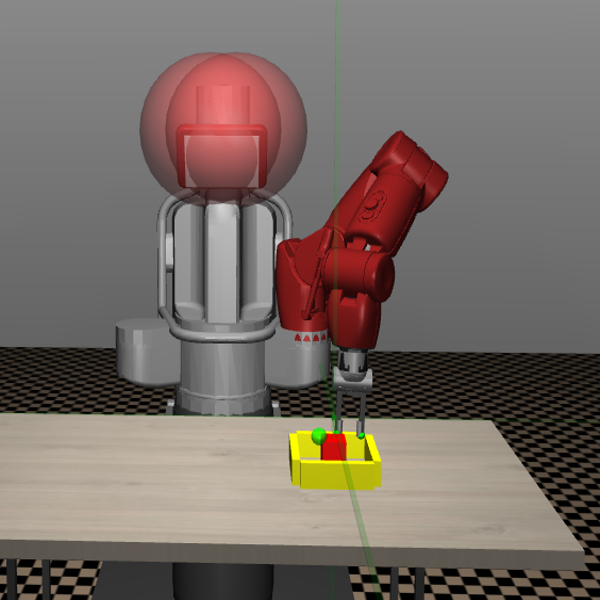}
  \caption{Take A out of B}
  \label{fig:sub4}
\end{subfigure}
\caption{Suite of robot manipulation tasks with baxter robot with end-effector control and parallel jaw gripper. }
\end{figure}

\begin{itemize}
    \item \textbf{Pick and Move} (Figure \ref{fig:sub1}) The robot needs to reach towards the object, grasp it and move to a target 3D location. 

\item \textbf{Put A inside B} (Figure \ref{fig:sub2}) The robot needs to reach towards the object, grasp it and put it inside the container. 

\item \textbf{Stack A on top of B} (Figure \ref{fig:sub3}) The robot needs to reach towards the red object, grasp it  and put it on top of the purple object. 

\item \textbf{Take A out of B} (Figure \ref{fig:sub4}) The robot needs to reach towards the object inside the container, grasp it and take it out of the container. The objective of this environment to grasp the objective in cluster container, then move out of container and move it to any 3D location in the workspace. 
\end{itemize}
All the tasks in our benchmark suite require long temporal horizon policies, which is a challenge when learning from  sparse  rewards.    

\begin{enumerate}
    \item How well the proposed exploration strategy performs over $\epsilon$-greedy?
    \item How out learnt policies compare with policies assembled directly  over macro-actions or options?
    \item What is the impact of the chosen skill set to the proposed method?
    \item What happens when the dynamics are quite different in the new environment?
\end{enumerate}

We evaluate our method against the following baselines: 
\begin{enumerate}
    \item Hindsight Experience Replay (\textit{HER}), described in \cite{andrychowicz2017hindsight}. In this method, exploration is carried out by sampling noise  from an $\epsilon$-normal  distribution and adding it to the predicted actions of the  policy learned so far. This is a state-of-the-art model-free RL method for control.
    \item Parameterized action space (\textit{PAS}), described in \cite{ICLR16-hausknecht}. This approach uses the learned skills as macro-actions by learning a meta-controller which predicts probability over those skills, as well as the goal of each skill. 
    \item Parameterized action space + Hierarchical HER (\textit{HER-PAS}), as described in  \cite{levy2017hierarchical}.  We extend the model of \cite{ICLR16-hausknecht} by creating additional hindsight experience at macro-level, i.e., in the parameterized action level (skill and sub-goal). Specifically, we replace the goal state in the macro-action transitions  collected during the episode which failed to achieve goal $\mathrm{g}$ with the $g' = s_T$ and evaluate all the low level actions, chosen using the skill and sub-goal, with the new reward function $r_t = r_{g_{\text{new}}}(s_t,a_t)$. 
    \item Parameter Space Noise + HER(\textit{HER+ParamNoise}), described in \cite{plappert2017parameter}, exploration is carried out in parameter space similar to evolutionary strategies. 
\end{enumerate}

\begin{figure}
\centering
\begin{subfigure}{.49\textwidth}
  \centering
  \includegraphics[height=5cm, width=\linewidth]{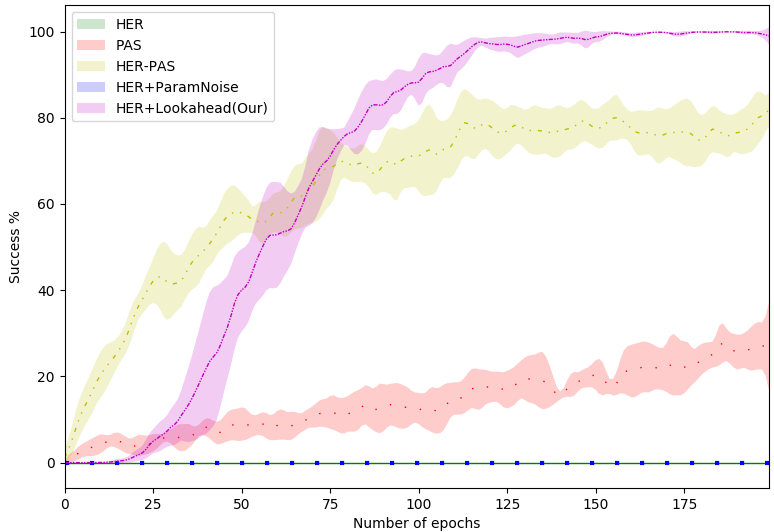}
  \caption{Pick and Move}
  \label{fig:succsub1}
\end{subfigure}
  \centering
\begin{subfigure}{.49\textwidth}
  \includegraphics[width=\linewidth]{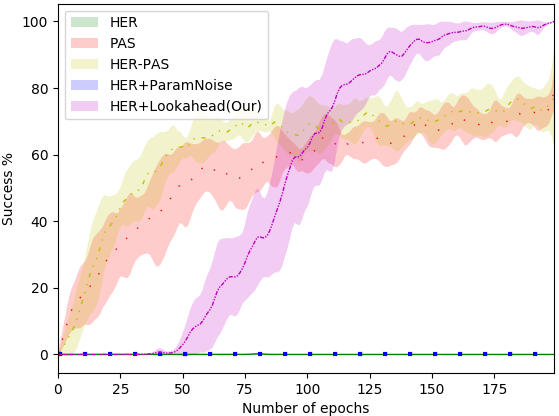}
  \caption{Put A inside B}
  \label{fig:succsub2}
\end{subfigure}

\begin{subfigure}{.49\textwidth}
  \centering
  \includegraphics[width=\linewidth]{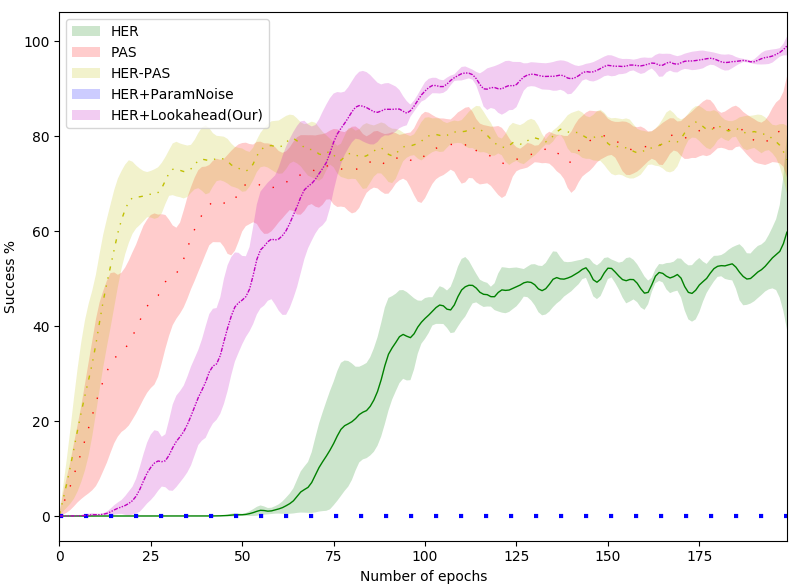}
  \caption{Stack A on top of B}
  \label{fig:succsub3}
\end{subfigure}
\begin{subfigure}{.49\textwidth}
  \centering
  \includegraphics[width=\linewidth]{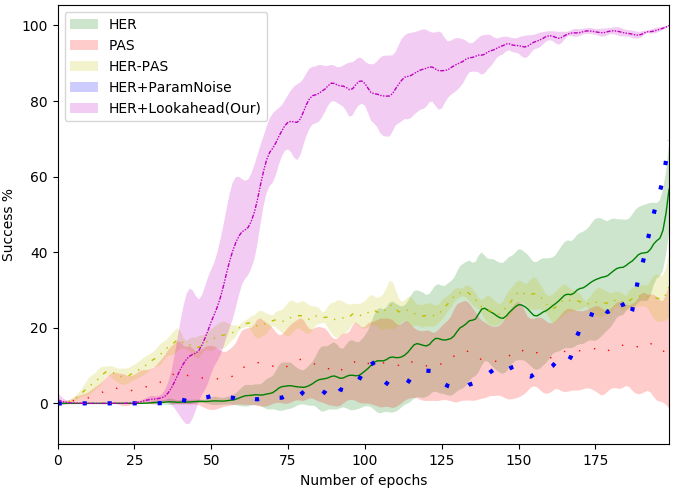}
  \caption{Take A out of B}
  \label{fig:succsub4}
\end{subfigure}
\caption{The success plot for each manipulation task in our suite. For evaluation, we freeze the current policy and sample 20 random initial states and goals at each epoch (1 epoch = 16 episodes of environment interaction).}
\label{fig:succplot}
\end{figure}

Success plots for the different tasks for our method and baselines are shown in  Figures \ref{fig:succplot}. Our method significantly outperforms the baselines in terms of final performance and sample complexity of environment interactions. 
In  ``put  A inside  B" task \ref{fig:sub2} \textit{HER}  is not able to succeed at all. PAS show success early in training but converges to sub-optimal performance.  This is expected   due to the restriction imposed on the policy space by to the hierarchical structure.

\begin{figure}
\centering
\begin{subfigure}{.49\textwidth}
  \centering
  \includegraphics[width=\linewidth]{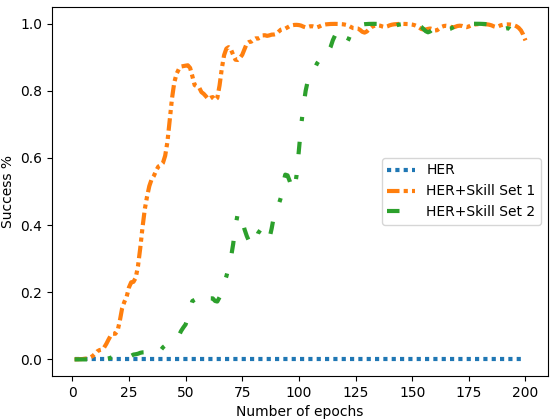}
  \caption{Pick and Move task}
  \label{fig:succsub1lib2}
\end{subfigure}
  \centering
\begin{subfigure}{.49\textwidth}
  \includegraphics[width=\linewidth]{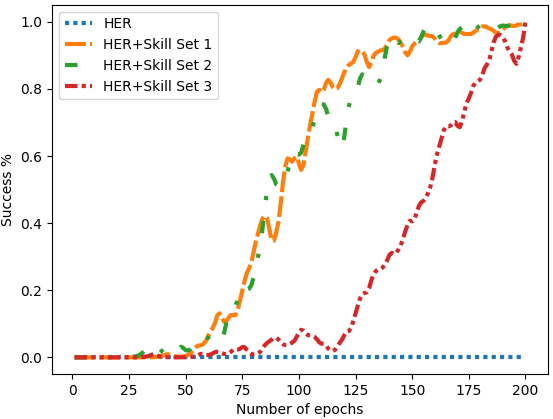}
  \caption{Put A inside B}
  \label{fig:succsub2lib2}
\end{subfigure}
\caption{\textbf{Sensitivity of our method to skill set selection.} We show the success plots for ``Pick and move"  and ``Put A inside B" tasks. Look-ahead exploration leads to slower learning when essential skills are missing. However, it is still better than random exploration.(1 epoch = 16 episodes of environment interaction).}
\label{fig:succplotlib2}
\end{figure}

\paragraph{Sensitivity to skill set selection} 
We use skills (macro-actions) for exploration, and not for assembling a hierarchical policy, as done in the option framework \cite{SUTTON1999181}. Therefore, we expect our method to be less sensitive to the skill selection, in comparison to previous methods \cite{SUTTON1999181}.
To quantify such flexibility we conducted experiments using three different skill sets: 
 faster convergence. To test this hypothesis, we used 3 skill set 
$\mathcal{K}_1$ = \{transit, grasp, transfer\}, $\mathcal{K}_2$ = \{grasp, transfer\} and $\mathcal{K}_3$ = \{transit, transfer\}. We  tested our method with $\mathcal{K}_2$ on the  ``Pick and Move", and ``Put A inside B" tasks and show results in Figure \ref{fig:succsub1lib2}. Exploring with all the skills ($\mathcal{K}_1$) leads to faster convergence of policy learning than using $\mathcal{K}_2$. This is easily explained as 
the agent needs to  learn the transit skill via interactions generated by other skills and the current policy. We did not observe any slower learning for ``Put A inside B" task using $\mathcal{K}_2$, as shown in Figure \ref{fig:succsub2lib2}. The reason for this  is that the transit skill is similar to the transfer skill, which is present.  Learning of ``Put A inside B" task using $\mathcal{K}_3$  is slower (Figure \ref{fig:succsub2lib2}).  This is due to the fact that grasping is a critical skill in completing this task. However our method is still able to learn the task from scratch, while \textit{HER}  fails to do so.

\begin{figure}[h]
\centering
\begin{subfigure}{.35\textwidth}
  \centering
  \includegraphics[width=\linewidth]{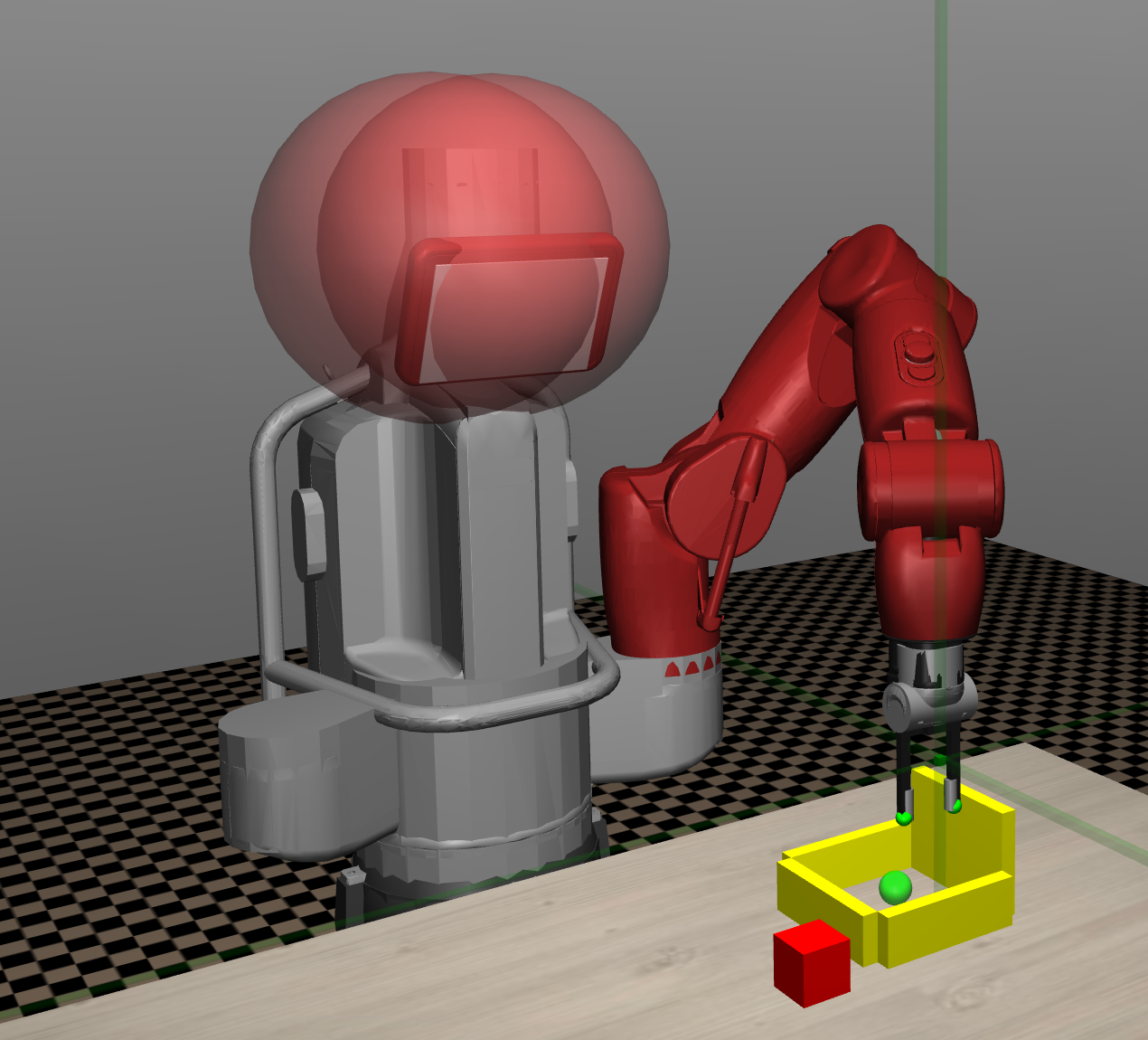}
  \caption{Put A inside B with 1 insurmountable wall}
  \label{fig:envhigh}
\end{subfigure}
  \centering
\begin{subfigure}{.55\textwidth}
  \includegraphics[width=\linewidth]{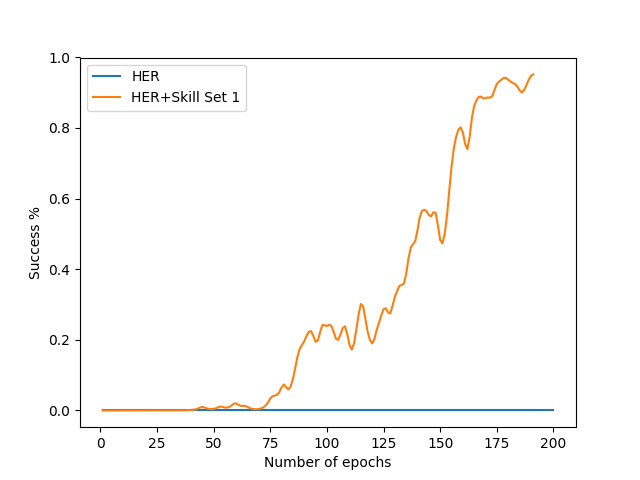}
  \caption{Success Curve}
  \label{fig:succhighwall}
\end{subfigure}

\caption{\textbf{Sensitivity of our method to model errors.} In (A) we show the perturbed version of ``Put A inside B" task in which a wall is very high and cannot be crossed by the agent. In (B) we show that even with wrong coarse dynamics model our approach works better than random exploration. (1 epoch = 16 episodes of environment interaction).}
\label{fig:succplotlib2high}
\end{figure}

\paragraph{Sensitivity to model errors} 
We  quantify sensitivity of our method to model errors, i.e., the accuracy of dynamics skill models.
We created a perturbation in our ``Put A inside B" task by making a wall insurmountable by the robot arm, shown in Figure \ref{fig:envhigh}. The skill dynamics that concern end-effector states near the large wall would be completely wrong, since this wall was absent during learning of our bacis skills and therefore their dynamics. However, the proposed look-ahead exploration is still beneficial, as seen in Figure \ref{fig:succhighwall}. Our method succeed to learn task, while \textit{HER} with random exploration  fails to do so. We used our full skill set in this experiment, namely, transit, transfer and grasp skills.

\textbf{Scalability}
Our method trades time of interacting with the environment with time to ``think" the next move by mentally unfolding a learned model of dynamics.  
The ``thinking" time depends on the speed of the look-ahead search.  
We experimented with different branching factors for the tree search.

With $\epsilon$-greedy the agent takes 0.4 seconds per episode (50 steps), with branching factor (bf) equal to 5 the agent takes 17 seconds, with bf=10 it takes 71 seconds, and with bf=15 it takes 286 seconds. However, for complex tasks in our benchmark suite, we did not observe empirical advantages from larger branching factors, so all the reported results use bf=5. 
The sampling process and pruning could be implemented in parallel on GPU, which will render our tree search much more efficient. We are currently exploring learning-based ways to guide the tree unfolding. 

\section{Conclusion - Future work}
We proposed an exploration method   that uses coarse-time dynamics of  basic manipulation skills for effective look-ahead exploration during learning of  manipulation policies. Our empirical findings suggest that the proposed  look-ahead exploration guided by learned dynamics of already mastered skills, can effectively reduce sample complexity when mastering a new task, to the extent that skills that are impossible with naive exploration, are possible with the proposed dynamics-based look-ahead exploration, suggesting an avenue for curriculum learning of manipulation policies, by continually expanding the skill set. The  resulting policies are still in the space of primitive low-level actions, which allows flexibility on choosing skills, and on the resulting reactive policy learner. 

If the proposed exploration strategies are used in environments that have  different dynamics than the environments used to train the basic skills (e.g., there is a wall present, or the object sizes are very different),  learning is slower, and our exploration scheme offers less advantages due to model errors. The fact that our dynamics are not updated online is a limitation of our method, and an avenue for future work. All the experiments performed in our paper are based on the centroid estimates of the objects of interest. It would be  interesting to explore policies and models learned directly from visual features in future work.

\section{Acknowledgment}
The authors would like to thank Chris Atkeson for useful discussions on learning coarse dynamics models.
This work was conducted in part through collaborative participation
in the Robotics Consortium sponsored by the U.S Army Research
Laboratory under the Collaborative Technology Alliance
Program, Cooperative Agreement W911NF-10-2-0016. The views
and conclusions contained in this document are those of the authors
and should not be interpreted as representing the official policies,
either expressed or implied, of the Army Research Laboratory of
the U.S. Government. The U.S. Government is authorized to reproduce
and distribute reprints for Government purposes not withstanding
any copyright notation herein.

\bibliography{egbibnew}
\clearpage
\appendix

\section{Environment Details}
The environments are similar to multi-goal environments proposed in \cite{plappert2018multi}.
We made this environment to make the results comparable with other works who use more famous OpenAI environments. Another reason for choosing baxter as our robot is due to its availability of platform in our lab for performing transfer to real world experiments which is left as a future work.

In all the environment, actions are 4-dimensional with first 3-dimension as the cartesian motion of the end-effector and last dimension controls the opening and closing of the gripper. We apply the same action 75 simulation steps(with $\Delta t=0.002$). The state space includes the cartesian position of the end-effector, its velocities, position and velocity of gripper, objects pose and its velocities.

We use a single starting state in which object is grasped.  The state space includes the relative position of object which has to be transported in then environment. The goal location is given as a 3D cartesian location where the object has to be transported to.

\subsection{Pick and move object to a target location in 3D}
This environment is a close replica of \cite{plappert2018multi} FetchPickAndPlace environment made for baxter robot. The objective to move the object from arbitrary location in the workspace to the 3D target location in the workspace.  The starting location of the gripper and object are sampled randomly in the workspace of the robot. The reward function is binary, i.e. the agent obtains reward 0 when the object is within the tolerance 3cm and -1 otherwise. 

\subsection{Put object A inside container B}
The objective is to go, grab the object from any arbitrary location in the workspace and move it inside the container. The goal is specified as the center of the container which is kept fixed across all the episodes. The binary reward function gives the agent 0 if the object is within 5cm of the target location and z of the object is less 2cm from the ground and -1 otherwise.

\subsection{Put object A on object B}
The objective is to go, grab the object from random location and put it over another object B, whose location is fixed. The goal is specified as the center of the top surface of the object B. The binary reward function gives the agent 0 if the object is within 3cm of the target location and object A is in contact of object B and -1 otherwise.

\subsection{Take object A out of container B}
The objective is to take reach to the object(random starting location inside the container), grab it and move it out of the container B. The goal is specified as a 3D location(randomly sampled) in the workspace of the robot. The agent gets reward 0 if the object is within 3cm of the target location and -1 otherwise.

\section{Policy learning for Skill Details}
We learnt policies for skills using Deep Deterministic Policy library with hindsight experience replay\cite{andrychowicz2017hindsight}. 
\subsection{Skills description}
We trained 3 skills namely transit, grasping and transfer object. The state space of transit is gripper location and goal is 3D target location for transit skill and action space is 3D movement of the end-effector. The state space and action space of grasping and transfer skill is same as PicknMove task. The starting configuration for grasping skill is gripper touching the object and goal is to raise object above ground. The starting configuration for transfer skill is object grasped in gripper and goal location is sampled randomly in the empty workspace of the robot.

\section{Successor Regresson model Details}
The coarse level successor predictor model is learnt for each skill individually after training each skill independently. 

\subsection{Data collection}
After training each skill, we sample starting configurations in the skill environments and let the trained skill act in its environment. We store the starting configuration and  the resulting state to act as target of the successor prediction model. For training the success predictor we sampled starting configuration from PicknMove environment.

\subsection{Model architecture}
For success probability prediction, we used fully connected network with 2 hidden layers of 50 and 100 neurons each. We trained the model with learning rate = 1e-3, binary cross-entropy and ADAM optimizer.

We used fully connected network with 3 hidden layers of 1000 neurons each for representing the successor prediction model. We trained the model with learning rate = 1e-5, l2 regression loss and ADAM optimizer.

\section{Implementation Details}
For representing actor and critic we used a neural network with 3 fully connected layers with 64 neurons each. For our methods and baselines we trained DDPG and build over \cite{baselines}. The other hyperparameters are as follows:
Actor Learning rate: 1e-3\\
Critic Learning rate: 1e-4\\
Target network update: 
gamma: 0.98\\
Number of cycles per epoch: 20\\
Number of updates per cycle: 40\\
Batch size: 128\\

\end{document}